\documentclass[letterpaper, 10 pt, conference]{ieeeconf}
\IEEEoverridecommandlockouts
\usepackage{amsmath,amssymb,amsfonts}
\usepackage{graphicx}
\usepackage{textcomp}
\usepackage[table]{xcolor}
\usepackage{booktabs}
\usepackage{makecell}
\usepackage{multirow}
\usepackage{caption}
\usepackage{subcaption}
\usepackage{algpseudocode}
\usepackage{algorithm}
\usepackage{float}
\usepackage{diagbox}
\usepackage{lineno}
\usepackage{setspace}
\usepackage{ulem}
\usepackage{xcolor}
\usepackage{moreverb,url}
\usepackage{amssymb}
\usepackage{lipsum}
\usepackage{amssymb}
\usepackage{mathtools}
\usepackage{array}
\usepackage{tabularx}

\title{\LARGE \bf
SteeredMarigold: Steering Diffusion Towards Depth Completion of Largely Incomplete Depth Maps
}

\author{Jakub Gregorek and Lazaros Nalpantidis%
\thanks{All authors are with the Department of Electrical and Photonics Engineering, DTU - Technical University of Denmark, Kgs. Lyngby, Denmark.
        {   \tt\footnotesize \{jagre,lanalpa\}@dtu.dk}}%
\thanks{This work has been funded and supported by the EU Horizon Europe project ``RoBétArmé” under the Grant Agreement 101058731.}%
}

\begin{document}

\maketitle
\thispagestyle{empty}
\pagestyle{empty}

\begin{abstract}

Even if the depth maps captured by RGB-D sensors deployed in real environments are often characterized by large areas missing valid depth measurements, the vast majority of depth completion methods still assumes depth values covering all areas of the scene. To address this limitation, we introduce SteeredMarigold, a training-free, zero-shot depth completion method capable of producing metric dense depth, even for largely incomplete depth maps. SteeredMarigold achieves this by using the available sparse depth points as conditions to steer a denoising diffusion probabilistic model. Our method outperforms relevant top-performing methods on the NYUv2 dataset, in tests where no depth was provided for a large area, achieving state-of-art performance and exhibiting remarkable robustness against depth map incompleteness.
Our source code is publicly available at https://steeredmarigold.github.io.
\end{abstract}

\section{Introduction}
\label{sec:intro}
\begin{figure}[t]
\medskip
\centering
    \includegraphics[width=0.49\linewidth]{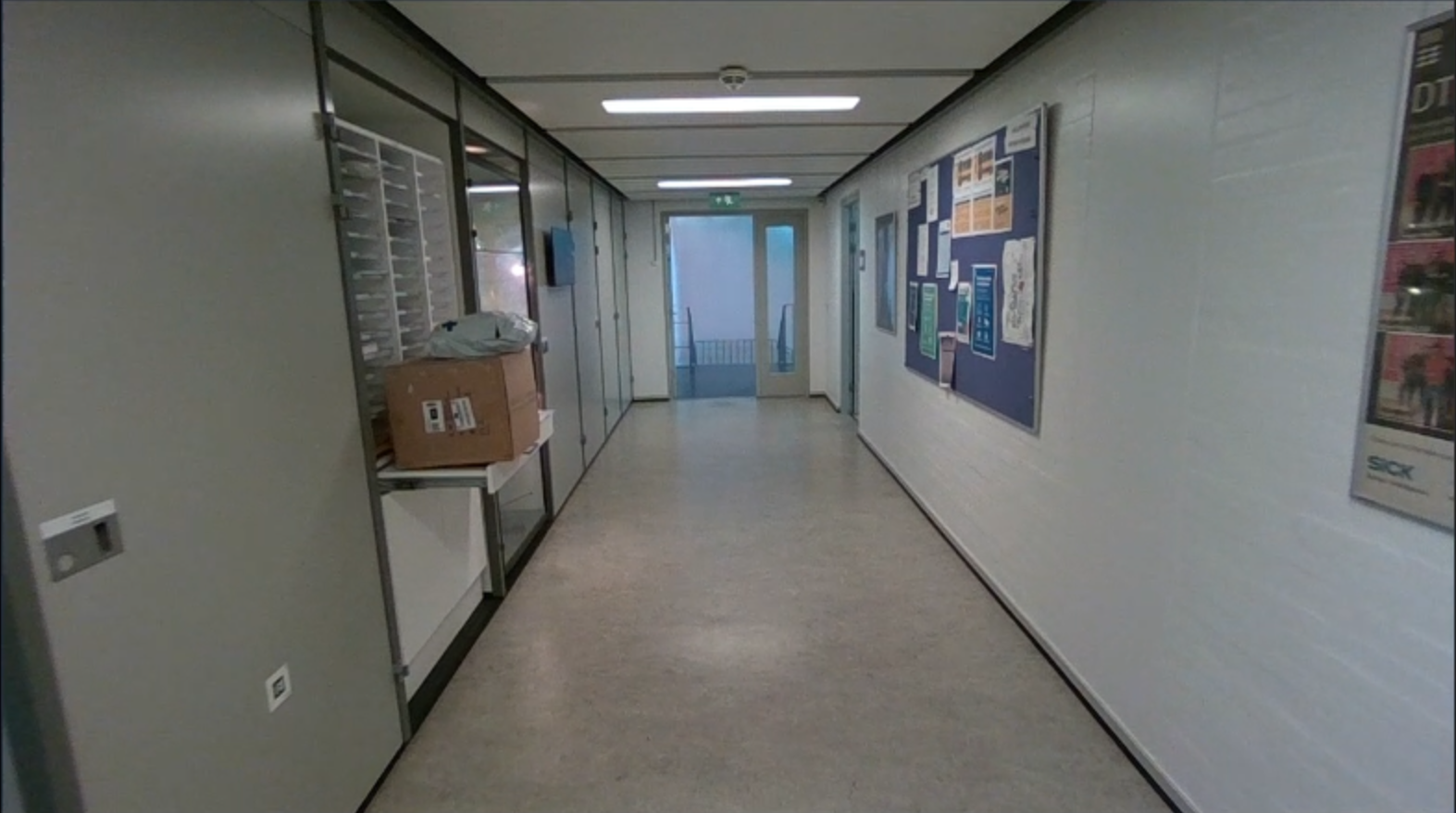}~
    \includegraphics[width=0.49\linewidth]{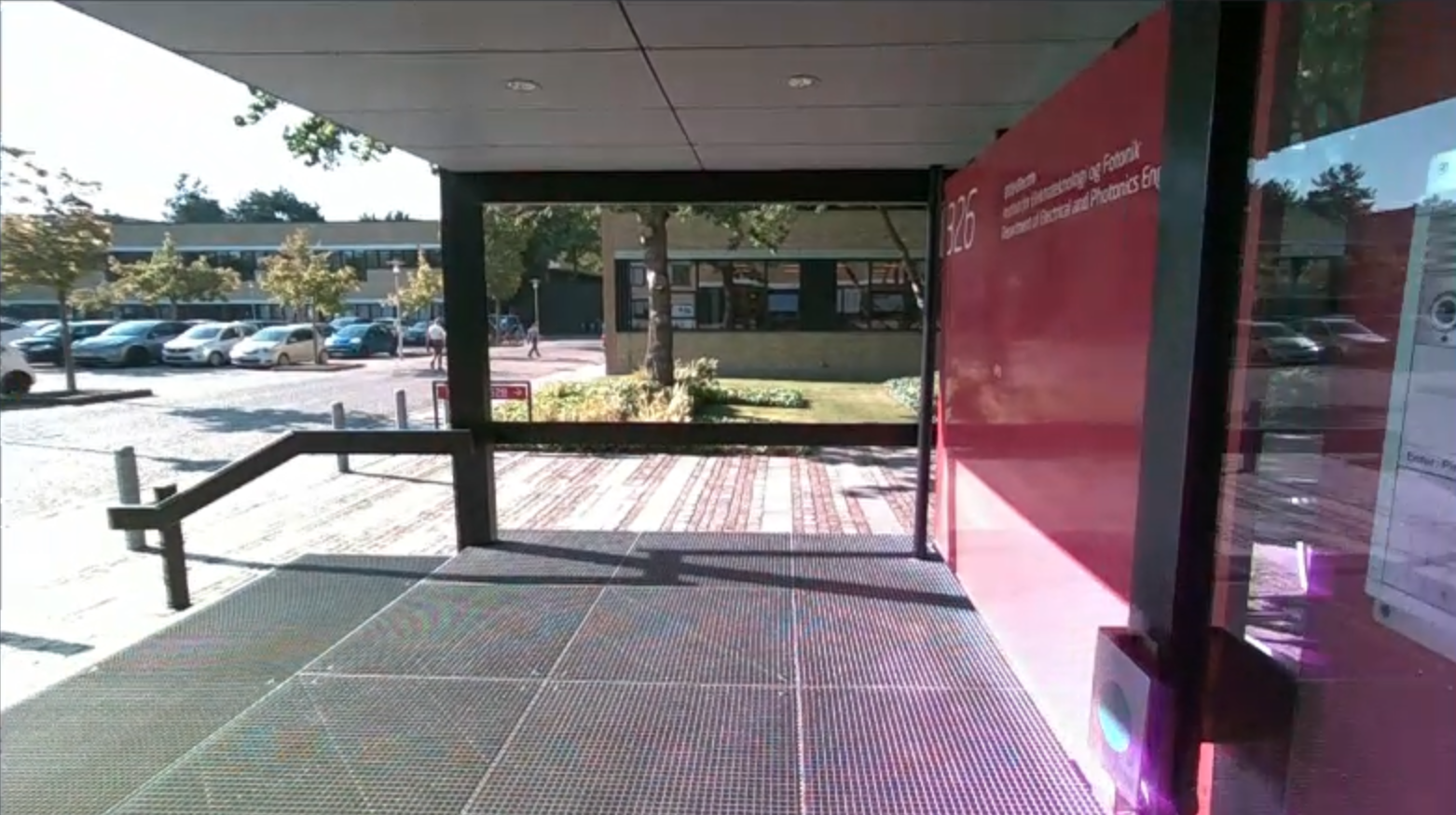}\\
    \smallskip
    \includegraphics[width=0.49\linewidth]{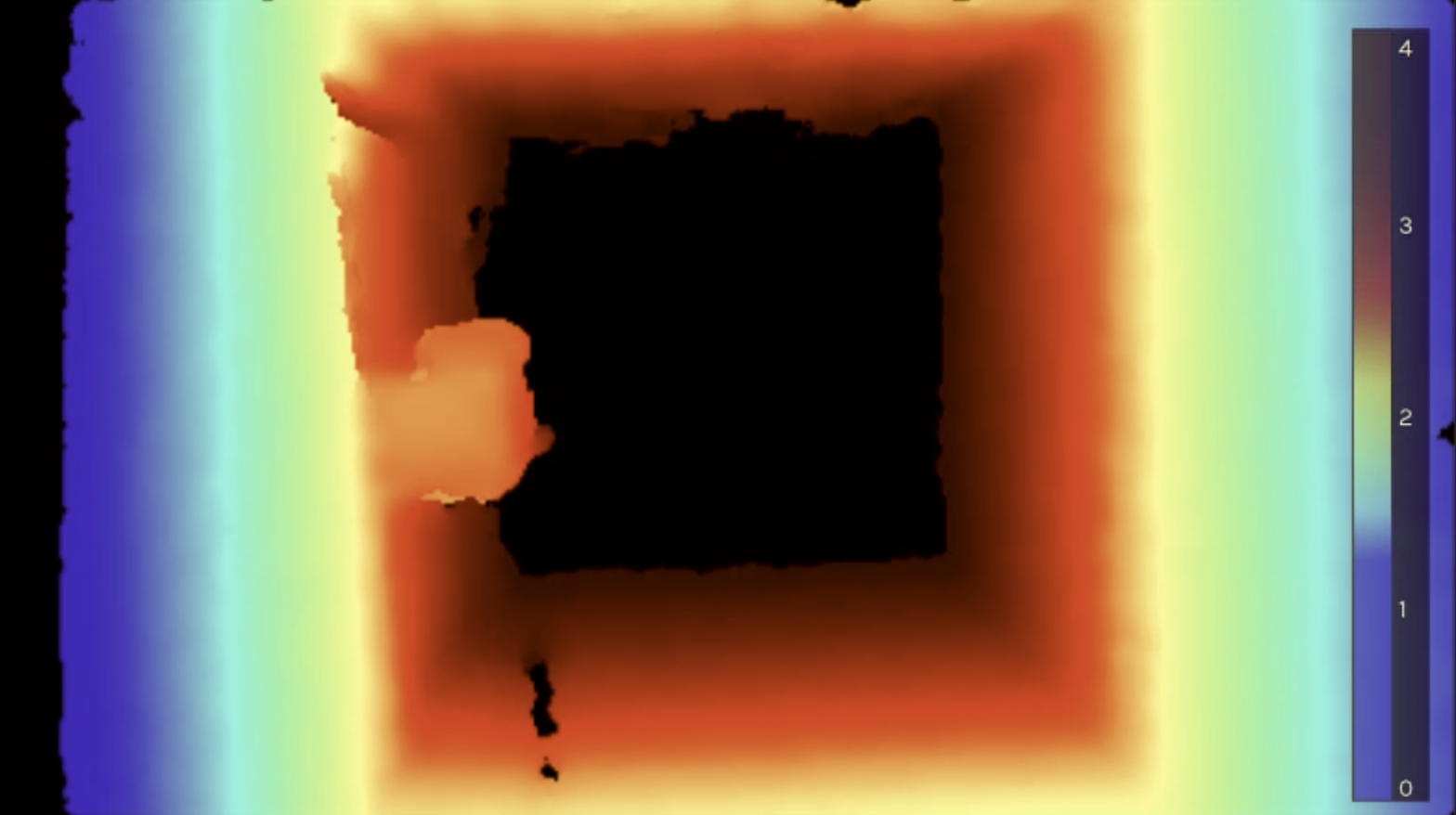}~
    \includegraphics[width=0.49\linewidth]{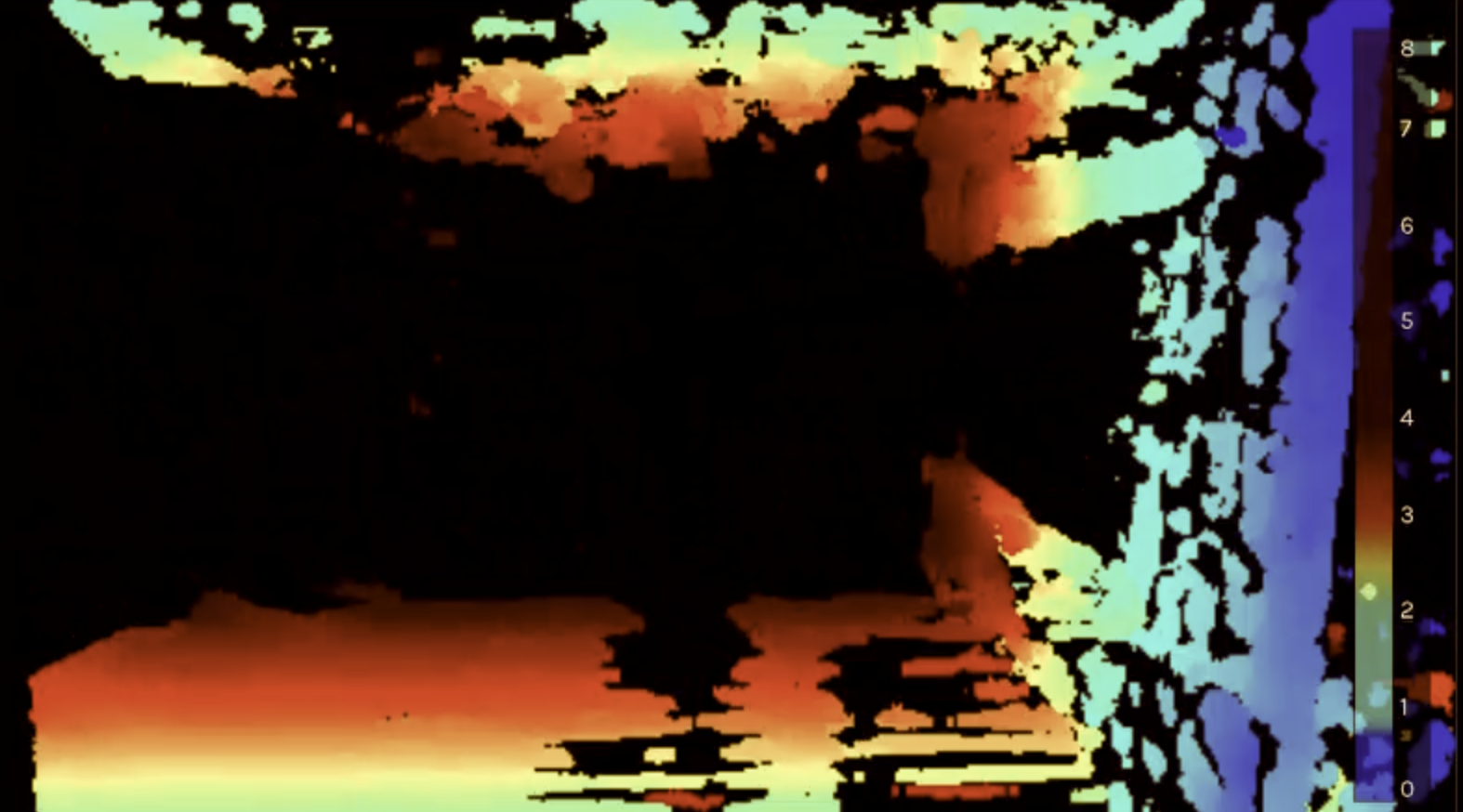}
\caption{RGB-D sensors often fail to provide robots with depth measurements in large areas due to large distances (case on the left) or lighting/material properties (case on the right). The task of completing such largely incomplete depth maps is much more challenging than the typical scenarios considered in the depth completion literature.}
\label{fig:teaser}
\end{figure}

This work focuses on providing robots with high-resolution, dense depth perception in cases of only partial depth measurements---possibly completely absent over large parts of the field of view. While numerous works have dealt with the closely related computer vision tasks of depth completion and monocular depth estimation, none of those approaches is used in robotics in practice. On the one hand, monocular depth estimation assumes no measured depth data present (which makes it too dangerous for robots), while on the other hand, depth completion assumes more or less uniformly-sparse depth values (which is very often violated by RGB-D sensors and sometimes even LiDARs). Actually, very few methods have focused on the real problem lying in-between: Depth completion of scenes with radically uneven depth sparsity \cite{zuo_ogni-dc_2024, conti_sparsity_2023, Arapis2022_ICRITA}, resulting in largely incomplete depth maps, such as e.g. the ones shown in Fig.~\ref{fig:teaser}.

We claim that this challenging task can be solved by performing depth estimation in empty regions---no matter how small or large---exploiting the available depth information in the periphery of the empty regions as constraints in the estimation process.
To this end, we have developed a novel model, dubbed \textit{SteeredMarigold}, that uses available sparse depth points as conditions to steer a denoising diffusion probabilistic model (DDPM)---we are using the Marigold model \cite{marigold}---to generate dense metric depth. The introduction of depth conditions to steer the diffusion process transforms a unimodal (RGB-only) depth estimator, like Marigold, into a strong depth completion model.

Our approach has several advantages that set it apart from current state-of-the-art depth completion methods. 
First, our SteeredMarigold model requires no training, as we capitalize upon the already trained Marigold model (which is utilized in a plug-and-play manner) and enhance its capabilities by depth conditioning. 
Second, our model inherits the universal nature of Marigold, which endows it with zero-shot depth completion capabilities in unseen environments. 
Third, the depth-steered diffusion process results in better metric and scale fidelity. This is attested by our model scoring significantly better than competing models without further retraining. Finally, our approach can be seen as a method for fusion of multiple modalities---i.e. RGB and depth, in the considered case of depth completion---but its fusion role could be easily generalized to other modalities and combinations. We make our code and model publicly available, hoping to attract more attention to the possibilities opened by the use of diffusion in multimodal perception tasks.

To summarize, the main contributions of this paper are as follows:
\begin{enumerate}
    \item We introduce \textit{SteeredMarigold}, a novel plug-and-play diffusion-based depth completion method capable of producing metrically- and scale-correct results.
    \item We formulate and address the challenging task of depth completion for scenes with radically uneven depth sparsity, achieving state-of-the-art performance.
\end{enumerate}

\section{Related Work}
\label{sec:related-work}

\textbf{Depth Completion.}
Depth completion is the task of calculating a dense depth map from an RGB image and a corresponding sparse depth map. In this task, Spatial Propagation Networks (SPN), initially proposed by Liu et al. \cite{liu_learning_2017}, achieve remarkable results. Notable extensions of SPN are CSPN \cite{cheng_depth_2018}, CSPN++ \cite{cheng_cspn_2020}, NLSPN \cite{park_non-local_2020} and DySPN \cite{lin_dyspn_2023}. SPN refinement modules were further improved in some of the latest models like BP-Net \cite{tang_bilateral_2024}, CompletionFormer \cite{Zhang_2023_CVPR} and PENet \cite{hu_penet_2021}.
While all those SPN-based approaches project the available depth points into the image plane, other approaches consider different projections, such as BEV@DC by Zhou et al. and the work of Yan et al. \cite{yan_tri-perspective_2024}. 
Other approaches take advantage of additional modalities: SemAttNet 
\cite{nazir_semattnet_2022} uses semantic labels as an additional information, whereas NDDepth \cite{shao_nddepth_2023} uses surface normals and plane-to-origin distance maps. 
A different approach was presented by Wang et al. \cite{wang_lrru_2023}, who used an architecture inspired by GuideNet \cite{tang2020learning} to refine a depth map produced by a fast classical image processing algorithm \cite{ku_defense_2018}.
In another work, Wang et al. \cite{wang_improving_2024} also proposed a method for guiding low-resolution depth features to high-resolution. Yan et al. \cite{yan_rignet_2023} proposed an architecture for feature extraction and depth generation based on dynamic convolutions, while Wang et al. \cite{wang_decomposed_2023}
proposed two decomposition schemes for guided dynamic filers, decreasing computational costs and memory footprint.

A major limitation of the aforementioned depth completion methods is that they are mostly trained on a single dataset and their performance drops dramatically when presented with data from other datasets. Furthermore, apart from the works of \cite{zuo_ogni-dc_2024, conti_sparsity_2023, Arapis2022_ICRITA} that consider varying depth sparsity, relevant literature generally assumes fixed and uniformly distributed depth coverage.

\textbf{Zero-Shot Depth Estimation.}
To mitigate specialization in specific datasets, universal models with zero-shot capabilities have been recently introduced: MiDaS \cite{midas} is the only method to
make predictions in the disparity space, DepthAnything \cite{depthanything} and Marigold \cite{marigold} predict relative, shift- and scale-invariant depth, while ZoeDepth \cite{zoedepth} and UniDepth \cite{piccinelli_unidepth_2024} are capable of
predicting metric depth. 
In contrast to \cite{midas, depthanything, zoedepth, piccinelli_unidepth_2024} that were trained using large quantities of data and computing capacity, \cite{marigold} benefits from large amounts of pre-trained knowledge of 
stable diffusion \cite{Rombach_2022_CVPR}, making its training on a single consumer GPU feasible. 
Ji et al. \cite{ddp} used a diffusion model for depth estimation as well as semantic segmentation. Duan et al. \cite{duan_diffusiondepth_2023} formulated self-diffusion learning, overcoming the need for dense depth ground truth. Both methods \cite{ddp, duan_diffusiondepth_2023} utilize pre-trained off-the-shelf Swin encoders \cite{liu_swin_2021}.

Even though these zero-shot depth estimation methods achieve impressive results, they struggle with metric and scale fidelity, while incorporating available depth measurements has not been yet possible.

\textbf{Diffusion Steering}
While diffusion models have proven themselves for depth estimation, fusion of multiple modalities---such as RGB and depth in the case of depth completion---remains out of their reach. Additional modalities could be introduced to diffusion models through channel concatenation like in \cite{marigold, ddp}, but this would require time- and compute-expensive retraining of the initial model. 
While conditioning models like ControlNet \cite{controlnet} or T2I-Adapter \cite{mou_t2i-adapter_2024} could be trained to introduce additional modalities, sparsity of depth data remains problematic. 
Alternatively, diffusion models can also be conditioned in a plug-and-play manner. Nair et al. \cite{nair_steered_2023} proposed a method providing fine-grained control over plug-and-play generation. 
A similar work was presented by Bansal et al. \cite{bansal_universal_2023}, who proposed a universal guidance mechanism utilizing estimation of clean data samples.
Another approach by Nair et al. \cite{nair_unite_2023} unites multiple pre-trained diffusion models---each introducing a different condition---leading to multi-modal image generation.
Other relevant approaches include \cite{lugmayr_repaint_2022, avrahami_blended_2022, kawar_imagic_2023, choi_ilvr_2021} and \cite{kawar_denoising_2022}. Finally, conditional generation by unconditional diffusion models in a plug-and-play manner was also explored by Graikos et al. \cite{graikos_diffusion_2022}.

Diffusion models have shown remarkable results in abstract tasks, such as inpainting, but none of the mentioned plug-and-play methods have been used in the context of metrically- and scale-strict tasks, such as depth completion.

\section{Method}
\label{sec:method}

Assume an RGB image $\mathbf{m} \in \mathbb{R}^{H \times W \times 3}$, the projection of sparse metric depth into the image plane $\mathbf{c}\in 
\mathbb{R}^{H \times W}$, and a 
dense metric depth prediction $\mathbf{d} \in \mathbb{R}^{H \times W}$. Our objective 
is to steer a diffusion process \textit{diff} in the latent space of a variational 
auto-encoder (VAE) to generate dense metric depth $\mathbf{d}$ respecting the sparse 
depth condition $\mathbf{c}$:
\begin{equation}
    \mathbf{d} = \mathcal{M}(\mathcal{D}(\textit{diff}(\mathcal{E}(\mathbf{m}), \mathbf{c})), \mathbf{c})
    \label{eq:the-first}
\end{equation}
where $\mathcal{E}$ is an encoder and $\mathcal{D}$ is a decoder. \textit{diff} is in our case
the depth estimator Marigold \cite{marigold} which predicts relative depth $\mathbf{d}^{*}$ that is transformed
to metric depth by a transformation $\mathcal{M}$:
\begin{equation}
    \mathbf{d} = \mathcal{M}(\mathbf{d}^{*}, \mathbf{c})
\end{equation}
using $\mathbf{c}$ as a reference. $\mathcal{M}$ performs least-squares fit to determine
scale and shift so as to transform the relative to metric depth.

We formulate our depth completion method as a plug-and-play conditioning for a pre-trained 
monocular depth estimator based on denoising diffusion probabilistic models (DDPM) \cite{ddpm}.

\subsection{Denoising Diffusion}
Diffusion models \cite{ddpm} are trained to reverse a forward noising
process fixed to a Markov chain, which adds Gaussian noise to a clean data sample 
$\mathbf{x}_{0}$ consecutively over a finite amount of time steps $0 < t \leq T$ 
until the signal is destroyed. The forward noising process is defined as:
\begin{equation}
   q(\mathbf{x}_{t}|\mathbf{x}_{t - 1}) \coloneq \mathcal{N}(\mathbf{x}_{t}; \sqrt{1 - \beta_{t}}\mathbf{x}_{t - 1}, \beta_{t}\mathbf{I})
   \label{eq:forward-process}
\end{equation}
where $\mathbf{x}_{1}, \dots, \mathbf{x}_{T}$ are latent variables, $\beta_{1}, \dots, \beta_{T}$ 
is a variance schedule, $\mathbf{I}$ is the identity matrix. The latent sample $\mathbf{x}_{t}$
at any given time point $t$ can be expressed as:
\begin{equation}
   \mathbf{x}_{t} = \sqrt{\bar{\alpha}_{t}}\mathbf{x}_{0} + \sqrt{1 - \bar{\alpha}_t} \epsilon
   \label{eq:forward-process-x0}
\end{equation}
where $\alpha_{t} \coloneq 1 - \beta_{t}$, $\bar{\alpha}_{t} \coloneq \prod_{s=1}^{t} \alpha_{s}$ and 
noise variable $\mathbf{\epsilon} \sim \mathcal{N}(0, \mathbf{I})$ of the same dimensionality as the
clean data sample and latent variables. The reverse process from time step $t$ to $t - 1$ is defined similarly:
\begin{equation}
   p_{\theta}(\mathbf{x}_{t - 1}|\mathbf{x}_{t}) \coloneq \mathcal{N}(\mathbf{x}_{t - 1}; \boldsymbol{\mu}_{\theta}(\mathbf{x}_{t}, t), \boldsymbol{\Sigma}_{\theta}(\mathbf{x}_{t}, t))
   \label{eq:reverse-process}
\end{equation}
where the mean $\boldsymbol{\mu}_{\theta}$ is expressed as:
\begin{equation}
    \boldsymbol{\mu}_{\theta} = 
        \frac{\sqrt{\bar{\alpha}_{t - 1}}\beta_t}{1 - \bar{\alpha}_t}\mathbf{x}_0 + 
        \frac{\sqrt{\alpha_t}(1 - \bar{\alpha}_{t - 1})}{1 - \bar{\alpha}_t}\mathbf{x}_t
\end{equation}
and the covariance $\boldsymbol{\Sigma}_{\theta} = \sigma^2_t \mathbf{I}$ is an untrained,
time-dependent constant where:
\begin{equation}
    \sigma_{t}^{2} = \frac{1 - \bar{\alpha}_{t-1}}{1 - \bar{\alpha}_t} \beta_t
\end{equation}
At any time step $t$ of the diffusion process, the clean data sample $\mathbf{x}_{0}$ can
be estimated as \cite{ddpm}:
\begin{equation}
   \mathbf{x}_{0} \approx \tilde{\mathbf{x}}_{0} 
      = \frac{\mathbf{x}_t - \sqrt{1 - \bar{\alpha}_{t}} \mathbf{\epsilon}_{\theta}(\mathbf{x}_{t})}{\sqrt{\bar{\alpha}_{t}}}
   \label{eq:x0-at-t}
\end{equation}
where $\mathbf{\epsilon}_{\theta}$ is a denoising neural network modeling 
$\epsilon$ given $\mathbf{x}_{t}$. If the model predicts velocity $\mathbf{v}$ \cite{salimans_progressive_2022} instead of the added noise, the clean data sample 
estimate is expressed as:
\begin{equation}
    \tilde{\mathbf{x}}_0 = 
        \sqrt{\bar{\alpha}_t}\mathbf{x}_t - \sqrt{1 - \bar{\alpha}_t}\mathbf{v}_\theta(\mathbf{x}_t)
    \label{eq:x0-at-t-v-objective}
\end{equation}

\subsection{Latent Space}

Following the Ke et al. \cite{marigold} we perform the diffusion in a lower dimensional latent 
space of VAE trained by Romach et al. \cite{Rombach_2022_CVPR}, which can be used to encode 
and decode depth with minimal error as:
\begin{equation}
    \mathbf{d}^{*} \approx avg(\mathcal{D}(\mathcal{E}(concat(\mathbf{d}^{*})))
\end{equation}
where $concat$ is a triple channel wise concatenation and $avg$ is channel-wise average.

\subsection{Steering}

In accordance with the ideas of \cite{nair_steered_2023, bansal_universal_2023, avrahami_blended_2022},
we use the estimate of clean data sample $\tilde{\mathbf{x}}_{0}$ to steer the diffusion process
and get closer to known depth measurements---sparse condition $\mathbf{c}$.

\begin{algorithm}[h]
    \caption{Steering DDPM towards depth completion}
    \label{alg:steering}
    \begin{algorithmic}
        \Require DDPM depth estimator, condition $\mathbf{c}$
        \State $\mathbf{x}_{T} \gets \mathcal{N}(0, \mathbf{I})$
        \For{$t = T \dots 1$}
            \State estimate clean data sample $\tilde{\mathbf{x}}_0$ (Eq. \ref{eq:x0-at-t-v-objective})
            \State perform reverse diffusion step, compute $\mathbf{x}_{t-1}$ (Eq. \ref{eq:reverse-process})
            \State shift $\mathbf{x}_{t-1}$ in the direction of $\mathbf{c}$ (Eq. \ref{eq:conditioning})
        \EndFor
        \State return $\mathbf{d}^{*} = avg(\mathcal{D}(\mathbf{x}_0))$
    \end{algorithmic}
\end{algorithm}
Algorithm \ref{alg:steering} illustrates how the steering is performed after each 
reverse diffusion time step using the following equation:
\begin{equation}
    \mathbf{x}_{t - 1} = \mathbf{x}_{t - 1} + \lambda \cdot (
        \mathcal{E}(\tilde{\mathbf{x}}^{\mathcal{D}}_0 - \phi_{1}(\tilde{\mathbf{x}}^{\mathcal{D}}_0, P) + 
        \phi_{2}(\tilde{\mathbf{x}}^{\mathcal{D}}_0, \mathbf{c}, P)) - \tilde{\mathbf{x}}_0
    )
    \label{eq:conditioning}
\end{equation}
where $\lambda$ is a steering factor controlling the steering strength, 
$\tilde{\mathbf{x}}^{\mathcal{D}}_0 = \mathcal{D}(\tilde{\mathbf{x}}_0)$ and $\phi_1$ and $\phi_2$ 
are functions performing linear interpolation of depth values sampled from 
$\tilde{\mathbf{x}}^{\mathcal{D}}_0$ exclusively or $\tilde{\mathbf{x}}^{\mathcal{D}}_0$ 
and $\mathbf{c}$ at positions $P$. The set of sampling positions $P$ is determined 
as a union of positions of all known depth values from $\mathbf{c}$ and randomly
selected locations for which the distance from the closest known depth value in 
$\mathbf{c}$ is larger than $\zeta$. $\phi_1$ interpolates depth values sampled at positions 
$P$ from $\tilde{\mathbf{x}}^{\mathcal{D}}_0$ while $\phi_2$ interpolates depth values 
sampled from $\mathbf{c}$ when a depth value at a particular position from $P$ is known 
in $\mathbf{c}$, otherwise the depth value is sampled from $\tilde{\mathbf{x}}^{\mathcal{D}}_0$.
Choi et al. \cite{choi_ilvr_2021} used low-pass filtering operation to replace high-level
features of the noisy sample in each diffusion step by the ones extracted from a given condition.
Inspired by this approach, we use linear interpolation to compute the linear component of 
$\tilde{\mathbf{x}}^{\mathcal{D}}_0$ and replace it by the one computed from $\mathbf{c}$.
This approach aims to preserve the fine-grained low-level features generated by
the diffusion model.

\begin{figure}[t]
    \smallskip
    \centering
    \includegraphics[width=1.00\linewidth]{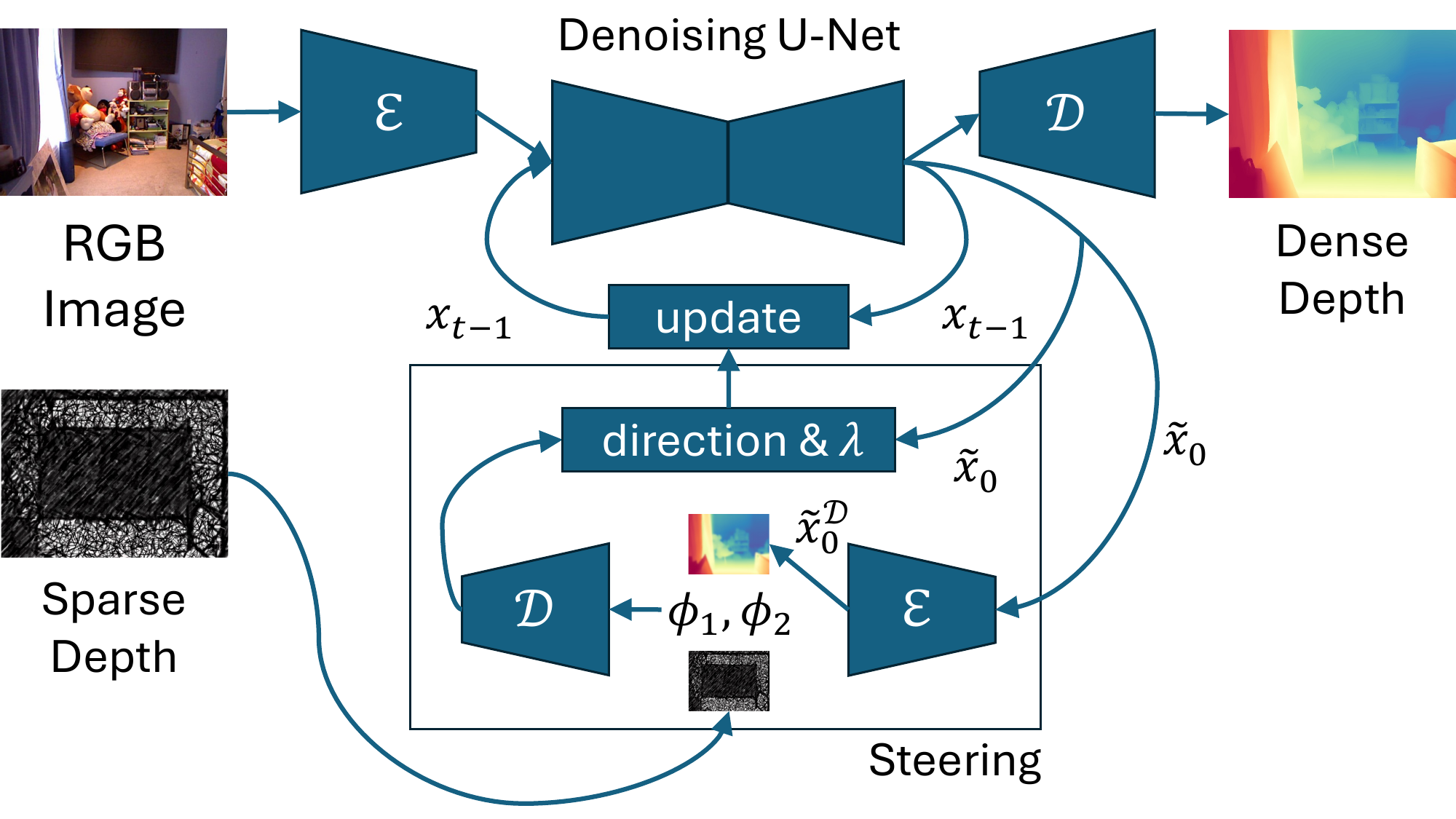}
    \caption{SteeredMarigold architecture. Our plug-and-play steering module expands the 
    Marigold diffusion model to perform depth completion. Note that there is only 
    one instance of the encoder $\mathcal{E}$ and decoder $\mathcal{D}$.}
    \label{fig:architecture}
\end{figure}

The condition $\mathbf{c}$ is assumed to be metric, while $\tilde{\mathbf{x}}^{\mathcal{D}}_0$
is relative. Thus, the depth values of $\mathbf{c}$ must be transformed to match the scale and shift
of $\tilde{\mathbf{x}}^{\mathcal{D}}_0$. This can be performed by 
$\mathcal{M}(\mathbf{c}, \tilde{\mathbf{x}}^{\mathcal{D}}_0)$.

Considering that we perform the diffusion in the VAE latent space and $\tilde{\mathbf{x}}_{0}$ is 
not directly interpretable, each steering step requires $\tilde{\mathbf{x}}_0$ to
be decoded to the image space and consequently its modified version must be encoded
back to the latent space.

The whole architecture of SteeredMarigold is visualized in Fig.~\ref{fig:architecture}.

\begin{figure*}[t]
\medskip
\centering
    \begin{subfigure}{.24\linewidth}
        \centering
        \includegraphics[width=1.0\linewidth]{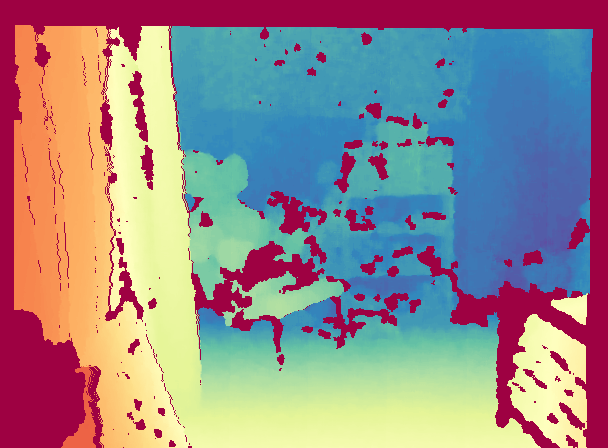}
        \caption{Ground-truth}
        \label{fig:available-ground-truth}
    \end{subfigure}
    \begin{subfigure}{.24\linewidth}
        \centering
        \includegraphics[width=1.0\linewidth]{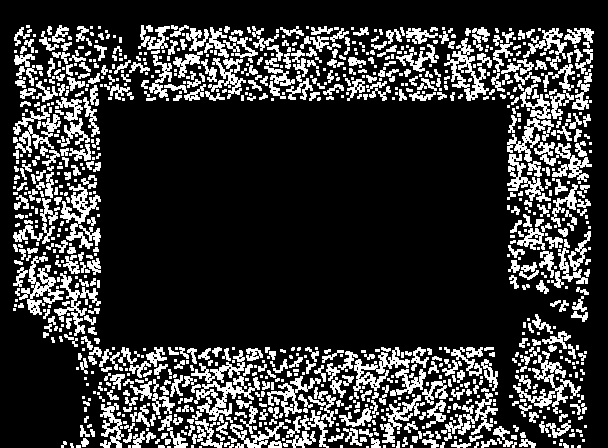}
        \caption{Sampled depth points}
        \label{fig:sampled-depth-missing-area}
    \end{subfigure}
    \begin{subfigure}{.24\linewidth}
        \centering
        \includegraphics[width=1.0\linewidth]{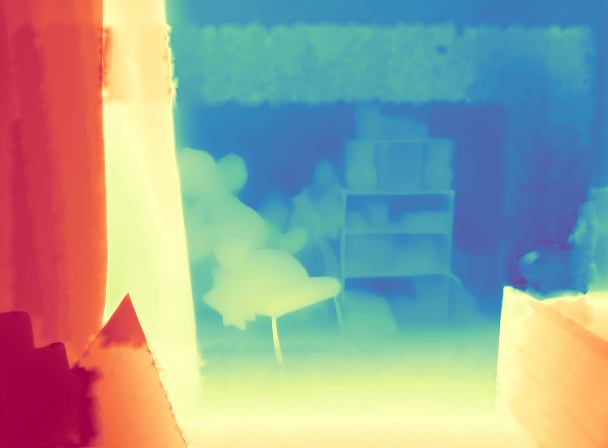}
        \caption{Initial steering}
        \label{fig:initial-steering}
    \end{subfigure}
    \begin{subfigure}{.24\linewidth}
        \centering
        \includegraphics[width=1.0\linewidth]{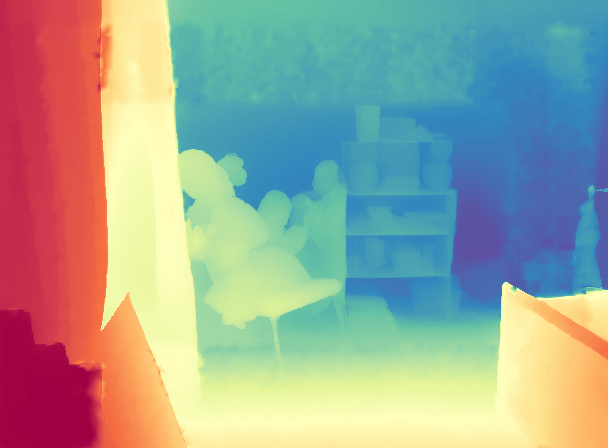}
        \caption{Final steering}
        \label{fig:later-steering}
    \end{subfigure}
    \caption{We can observe how the diffusion process is able to harmonize the depth estimate
    with the direction of steering. Depth points sampled from ground-truth (a)
    using the mask (b) are used to steer the diffusion process in the direction determined 
    by modifying $\tilde{\mathbf{x}}^{\mathcal{D}}_{0}$ using $\phi_1$ and $\phi_2$. 
    The effects of steering are very apparent in the initial steps (c) of the diffusion 
    reverse process and progressively becomes less visible in the latter steps (d) as the 
    diffusion process progressively harmonizes regions not affected by steering with the
    regions that are being impacted by the steering. The steering direction in (c) and (d)
    corresponds to $\tilde{\mathbf{x}}^{\mathcal{D}}_{0}$ after subtracting $\phi_1$ and 
    adding $\phi_2$. The visualization does not take the steering factor $\lambda$ into account.}
    \label{fig:steering-missing-area}
\end{figure*}

\section{Experimental Setup}
\label{sec:experimental-setup}

\textbf{Implementation.}
We implemented our method using PyTorch and the Diffusers library \cite{huggingface-diffusers}.
The Marigold depth estimator weights were retrieved from \cite{huggingface-marigold}. The diffusion 
steering is performed in a plug-and-play manner and does not require any training.

\textbf{Datasets.}
We evaluated our method on the indoor dataset NYUv2 \cite{Silberman:ECCV12}. NYUv2 consists of 464 scenes captured 
by a Kinect RGB-D sensor in resolution $640\times480$ pixels. We utilized only the testing
set containing 654 images.

\textbf{Metrics.}
Combining the standard practices from depth estimation and depth completion literature 
we report metrics that cover both tasks. More precisely, we report root mean square error 
$RMSE = \sqrt{\frac{1}{N} \sum_{i} \lvert \mathbf{d}_i - \mathbf{g}_i \rvert^{2}}$,
mean absolute error $MAE = \frac{1}{N} \sum_{i} \lvert \mathbf{d}_i - \mathbf{g}_i \rvert$,
mean absolute relative error $REL = \frac{1}{N}\sum_i\lvert\frac{\mathbf{d}_i - 
\mathbf{g}_i}{\mathbf{g}_i}\rvert$, and percentage of $i$ such that $\delta_1 = 
\max\left(\frac{\mathbf{d}_i}{\mathbf{g}_i},\frac{\mathbf{g}_i}{\mathbf{d}_i}\right) 
< 1.25$. $N$ denotes count of pixels and $\mathbf{d}_i$ and $\mathbf{g}_i$ are pixels 
of depth prediction and ground-truth. In this paper we express $RMSE$, $MAE$ and $REL$ 
in meters and $\delta_1$ in $\%\cdot10^{-2}$.

\begin{figure}[h]
    \centering
    \includegraphics[width=0.7\linewidth]{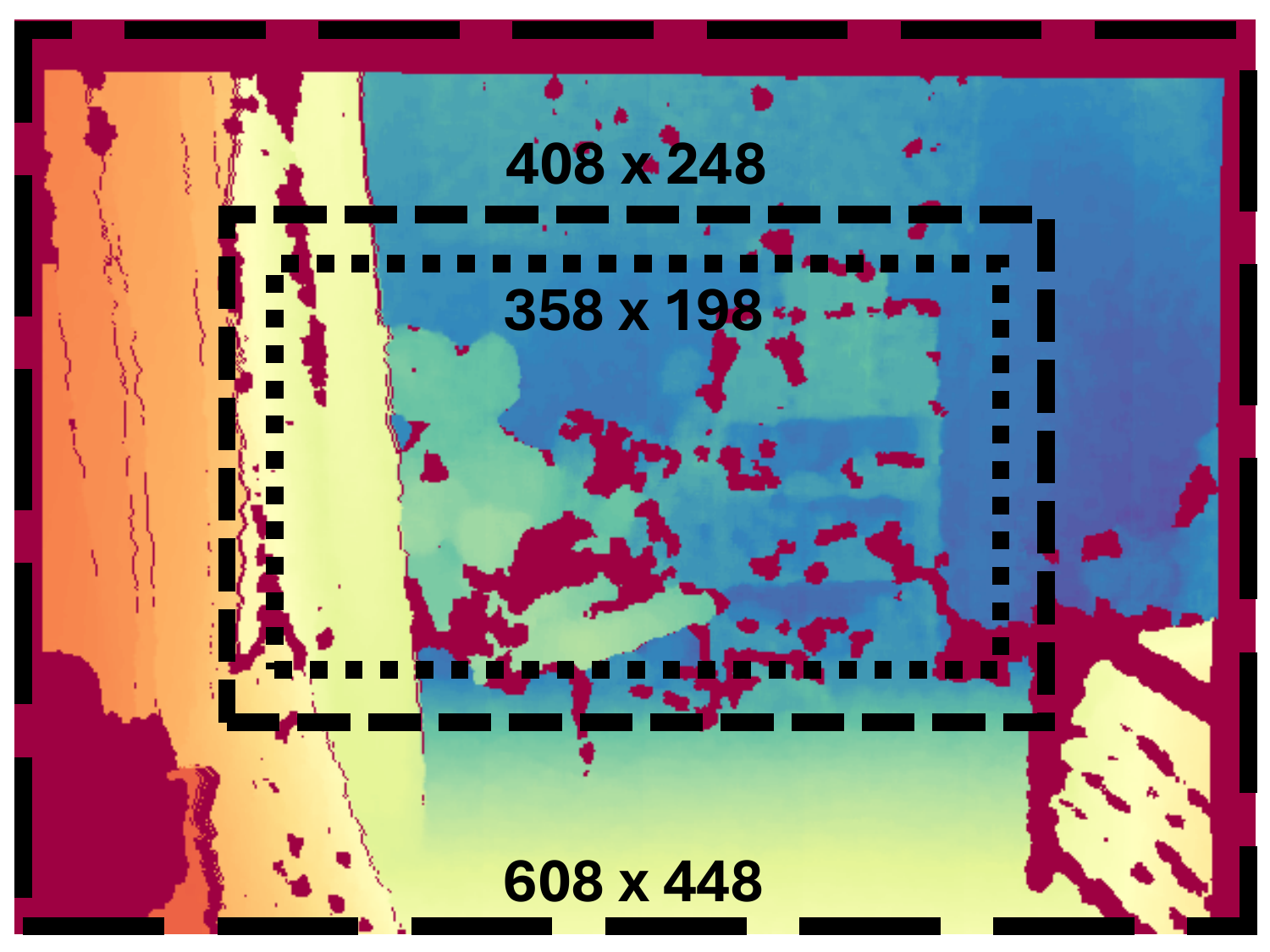}
    \caption{The three considered evaluation areas: large area ($608\times448$ - entire image), 
    medium area ($408\times248$ - equal to removed depth area) and small area ($358\times198$).}
    \label{fig:evaluation-areas}
\end{figure}

\textbf{Evaluation Protocol.}
The common evaluation practices on NYUv2 dataset are to down-sample the images by half
to resolution $320 \times 240$ and center-crop them to $304 \times 228$ pixels. Models
are then presented by 500 randomly sampled depth points. Being interested in applications 
requiring higher resolutions, we decided to evaluate our method in higher resolution 
$608\times448$ skipping the down-sampling and slightly increasing the cropping making
the resolution compatible with the selected baseline models as well as our method.
The higher resolution comes with the challenge of sparsity of the depth data. 
Even though this work touches upon different sparsity levels (similar to \cite{zuo_ogni-dc_2024}),
we mainly explore depth completion in situations where data from the depth sensor 
is not available for large parts of the scene (like illustrated in the \ref{fig:teaser}).
For this purpose we define three evaluation areas visualized in the Fig.~\ref{fig:evaluation-areas}.
Our approach is also different from \cite{conti_sparsity_2023} where a variety of sparsity 
patterns were considered.

We did not employ the ensembling scheme used with Marigold \cite{marigold} and limit our
method to a single diffusion run in 50 steps according to DDPM \cite{ddpm}.

\section{Results}
\label{sec:results}

\begin{table*}[h]
    \smallskip
    \caption{Performance on NYUv2 for selected baseline models, Marigold and SteeredMarigold under various 
    sparsity levels in the increased image resolution of $608\times448$. An $*$ denotes results reported 
    by authors of the original papers in the resolution and sampling commonly used for evaluation on 
    NYUv2 dataset.}
    \label{tab:best-results}
    \begin{center}\small{
        \begin{tabular}{cccccccc}
\toprule
Method                                 & Evaluation Area           & \# Depth & $\lambda$                       & REL $\downarrow$ & $\delta_1 \uparrow$ & RMSE $\downarrow$ & MAE $\downarrow$ \\ 
\midrule
BP-Net\textsuperscript{$*$}            & $304\times228$            & 500      & -                               & 0.012            & 0.996               & 0.089             & -                \\
\noalign{\smallskip}\cline{1-8}\noalign{\smallskip}
\multirow{3}{*}{BP-Net}                & $608\times448$            & 500      & -                               & 0.0619           & 0.9228              & 0.3019            & 0.1562           \\ 
                                       & $608\times448$            & 2000     & -                               & 0.0114           & 0.9911              & 0.1182            & 0.0308           \\ 
                                       & $608\times448$            & 13620    & -                               & 0.0066           & 0.9955              & 0.0862            & 0.0190           \\ 
\noalign{\smallskip}\cline{1-8}\noalign{\smallskip}
CompletionFormer\textsuperscript{$*$}  & $304\times228$            & 500      & -                               & 0.012            & -                   & 0.090             & -                \\
\noalign{\smallskip}\cline{1-8}\noalign{\smallskip}
\multirow{3}{*}{CompletionFormer}      & $608\times448$            & 500      & -                               & 0.0452           & 0.9550              & 0.2570            & 0.1297           \\
                                       & $608\times448$            & 2000     & -                               & 0.0072           & 0.9974              & 0.0725            & 0.0234           \\
                                       & $608\times448$            & 13620    & -                               & \textbf{0.0043}  & \textbf{0.9988}     & \textbf{0.0506}   & \textbf{0.0147}  \\
\noalign{\smallskip}\cline{1-8}\noalign{\smallskip}                        
Marigold\textsuperscript               & $608\times448$            & 13620    & -                               & 0.0615           & 0.9548              & 0.2478            & 0.1555           \\
\noalign{\smallskip}\cline{1-8}\noalign{\smallskip}                        
\multirow{4}{*}{SteeredMarigold (Ours)}& $608\times448$            & 13620    & $0.1 \sqrt{1 - \bar{\alpha}}_t$ & 0.0196           & 0.9965              & 0.0969            & 0.0524           \\
                                       & $608\times448$            & 13620    & $0.2 \sqrt{1 - \bar{\alpha}}_t$ & 0.0144           & 0.9979              & 0.0777            & 0.0392           \\ 
                                       & $608\times448$            & 13620    & $0.3 \sqrt{1 - \bar{\alpha}}_t$ & 0.0124           & 0.9982              & 0.0714            & 0.0342           \\
                                       & $608\times448$            & 13620    & $0.4 \sqrt{1 - \bar{\alpha}}_t$ & 0.0114           & 0.9983              & 0.0687            & 0.0317           \\
\bottomrule
        \end{tabular}}
    \end{center}
\end{table*}

\begin{table*}[h]
    \caption{Models evaluated on NYUv2 with depth samples erased from the centrally located region of 
    $408\times248$ pixels (medium area depicted in the Fig.~\ref{fig:evaluation-areas}). 
    The performance of BP-Net and CompletionFormer significantly drops compared to full depth 
    sampling reported in Table~\ref{tab:best-results}. The baseline methods struggle to 
    predict the depth in the erased area with the exception of Marigold and SteeredMarigold.
    This can be visually also observed in the Fig.~\ref{fig:full-depth-vs-ma}.
    }
    \label{tab:best-results-missing-area}
    \begin{center}\small{
        \begin{tabular}{cccccccc}
\toprule
Method                                          & Evaluation Area & \# Depth                    & $\lambda$                       & REL $\downarrow$    & $\delta_1 \uparrow$ & RMSE $\downarrow$ & MAE $\downarrow$ \\ 
\midrule
\multirow{9}{*}{BP-Net}                         & $608\times448$  & $500 - \text{erased}$       & -                               & 0.1875              & 0.5969              & 0.8575            & 0.5590           \\ 
                                                & $608\times448$  & $2000 - \text{erased}$      & -                               & 0.1467              & 0.6644              & 0.8068            & 0.4547           \\
                                                & $608\times448$  & $13620 - \text{erased}$     & -                               & 0.1358              & 0.6788              & 0.7700            & 0.4222           \\
                                                & $408\times248$  & $500 - \text{erased}$       & -                               & 0.3574              & 0.1374              & 1.2973            & 1.1343           \\ 
                                                & $408\times248$  & $2000 - \text{erased}$      & -                               & 0.3542              & 0.1440              & 1.2770            & 1.1135           \\
                                                & $408\times248$  & $13620 - \text{erased}$     & -                               & 0.3426              & 0.1627              & 1.2335            & 1.0728           \\
                                                & $358\times198$  & $500 - \text{erased}$       & -                               & 0.3817              & 0.1413              & 1.3950            & 1.2341           \\
                                                & $358\times198$  & $2000 - \text{erased}$      & -                               & 0.3874              & 0.1238              & 1.3966            & 1.2427           \\
                                                & $358\times198$  & $13620 - \text{erased}$     & -                               & 0.3790              & 0.1234              & 1.3602            & 1.2140           \\
\noalign{\smallskip}\cline{1-8}\noalign{\smallskip}
\multirow{9}{*}{CompletionFormer}               & $608\times448$  & $500 - \text{erased}$       & -                               & 0.1621              & 0.6575              & 0.7879            & 0.5115           \\
                                                & $608\times448$  & $2000 - \text{erased}$      & -                               & 0.1095              & 0.7638              & 0.6496            & 0.3541           \\
                                                & $608\times448$  & $13620 - \text{erased}$     & -                               & 0.0987              & 0.7857              & 0.6107            & 0.3200           \\
                                                & $408\times248$  & $500 - \text{erased}$       & -                               & 0.2770              & 0.3793              & 1.1325            & 0.9365           \\
                                                & $408\times248$  & $2000 - \text{erased}$      & -                               & 0.2502              & 0.4409              & 1.0027            & 0.8167           \\
                                                & $408\times248$  & $13620 - \text{erased}$     & -                               & 0.2311              & 0.4877              & 0.9472            & 0.7546           \\
                                                & $358\times198$  & $500 - \text{erased}$       & -                               & 0.2965              & 0.3658              & 1.2151            & 1.0247           \\
                                                & $358\times198$  & $2000 - \text{erased}$      & -                               & 0.2691              & 0.4299              & 1.0799            & 0.9023           \\
                                                & $358\times198$  & $13620 - \text{erased}$     & -                               & 0.2575              & 0.4511              & 1.0284            & 0.8563           \\
\noalign{\smallskip}\cline{1-8}\noalign{\smallskip}
\multirow{3}{*}{Marigold}                       & $608\times448$  & $13620 - \text{erased}$     & -                               & 0.0618              & 0.9559              & 0.2555            & 0.1599            \\ 
                                                & $408\times248$  & $13620 - \text{erased}$     & -                               & 0.0610              & 0.9570              & 0.2794            & 0.1799            \\
                                                & $358\times198$  & $13620 - \text{erased}$     & -                               & 0.0630              & 0.9536              & 0.2906            & 0.1912            \\
\noalign{\smallskip}\cline{1-8}\noalign{\smallskip}
\multirow{3}{*}{SteeredMarigold (Ours)}         & $608\times448$  & $13620 - \text{erased}$     & $0.1 \sqrt{1 - \bar{\alpha}}_t$ & \textbf{0.0352}     & \textbf{0.9834}     & \textbf{0.1854}   & \textbf{0.0960}  \\
                                                & $408\times248$  & $13620 - \text{erased}$     & $0.1 \sqrt{1 - \bar{\alpha}}_t$ & 0.0510              & 0.9718              & 0.2586           & 0.1523           \\
                                                & $358\times198$  & $13620 - \text{erased}$     & $0.1 \sqrt{1 - \bar{\alpha}}_t$ & 0.0573              & 0.9646              & 0.2850           & 0.1743           \\
\bottomrule
        \end{tabular}}
    \end{center}
\end{table*}

\begin{figure}[t]
    \medskip
    \centering
    \begin{subfigure}{0.32\linewidth}
        \centering
        \includegraphics[width=1.0\linewidth]{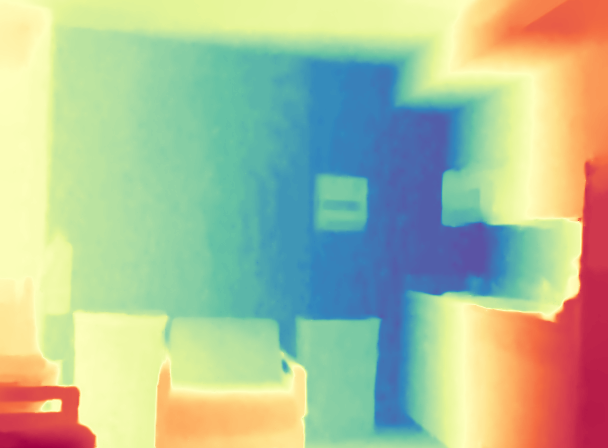}\\
        \smallskip
        \includegraphics[width=1.0\linewidth]{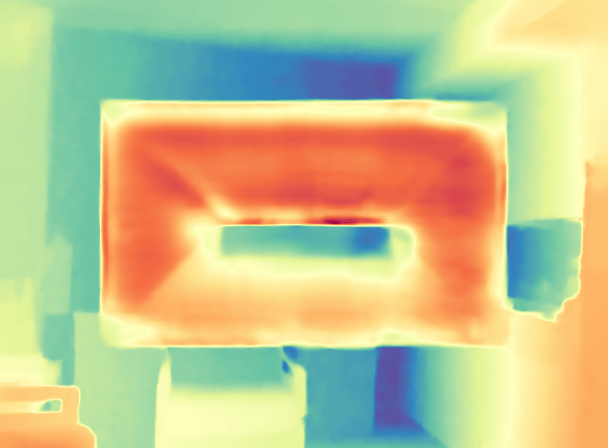}
        \caption{BP-Net}
        \label{fig:full-depth-vs-ma-bpnet}
    \end{subfigure}
    \begin{subfigure}{0.32\linewidth}
        \centering
        \includegraphics[width=1.0\linewidth]{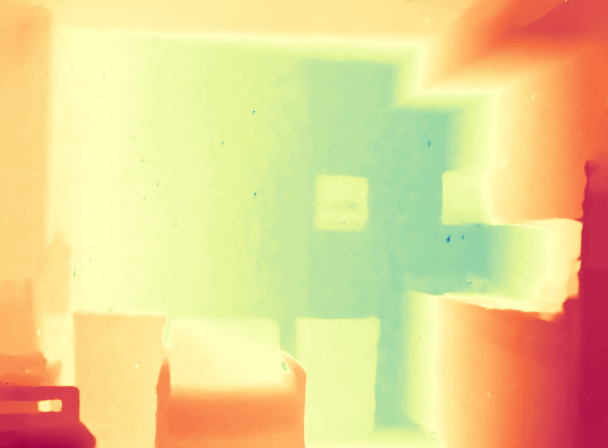}\\
        \smallskip
        \includegraphics[width=1.0\linewidth]{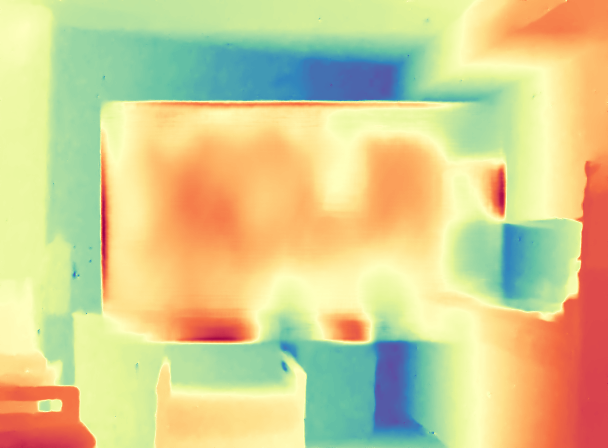}
        \caption{CFormer}
        \label{fig:full-depth-vs-ma-completionformer}
    \end{subfigure}
    \begin{subfigure}{0.32\linewidth}
        \centering
        \includegraphics[width=1.0\linewidth]{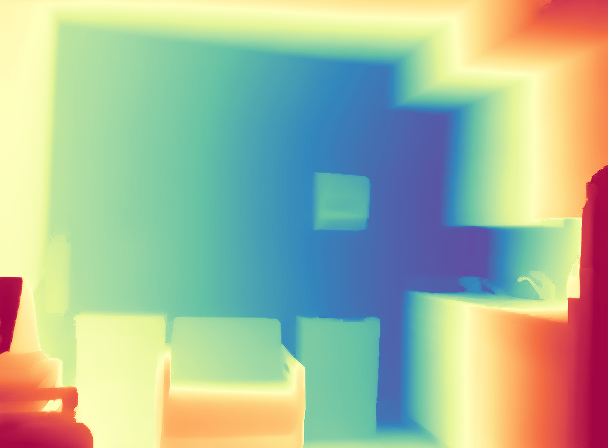}\\
        \smallskip
        \includegraphics[width=1.0\linewidth]{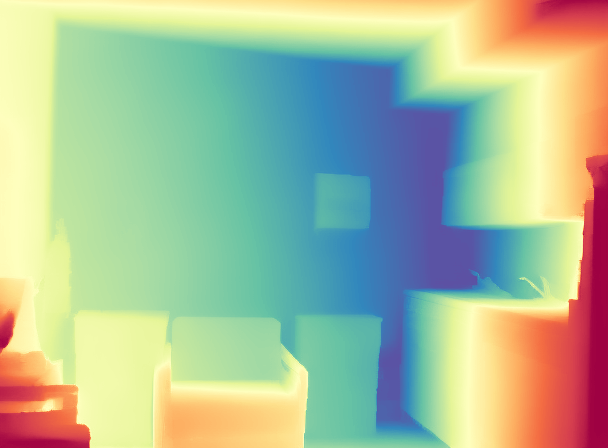}
        \caption{Ours}
        \label{fig:full-depth-vs-ma-steeredmarigold}
    \end{subfigure}
    \caption{Visualization of a completed scene by BP-Net (a), 
    CompletionFormer (b) and our method 
    (c). In the top row, the models were provided with 
    depth samples covering the entire scene. In the second row, no depth samples were provided in the central area of 
    $408\times248$. We can observe, that BP-Net
    and CompletionFormer struggle to complete the scene without
    any depth values in that region.}
    \label{fig:full-depth-vs-ma}
\end{figure}

We selected two models, BP-Net \cite{tang_bilateral_2024} and CompletionFormer \cite{Zhang_2023_CVPR}
as baselines. Both models are amongst the top performers on NYUv2 dataset \cite{Silberman:ECCV12} 
(they are also amongst the top performers on KITTI DC \cite{kitti_depth_dataset} dataset, which
we did not use for evaluation in this paper but inspired our evaluation). We benchmarked the mentioned models together with \textit{Marigold}
and our method \textit{SteeredMarigold}. It is important to note, that both baseline models were 
benchmarked with the weights published by their authors. No additional training was performed.
Three levels of depth sparsity were considered: (i) 500 depth samples, matching the standard 
practice of evaluating on the NYUv2 dataset; (ii) 2000 depth samples to approximate 
the depth sparsity the models were originally trained for (our evaluation resolution is approximately 
4 times higher than the models were trained for); and (iii) 13620 points which corresponds to the level of 
sparsity of LiDAR $~5\%$ points in the KITTI DC \cite{kitti_depth_dataset} dataset. 

Our results are presented in Table~\ref{tab:best-results}, but for completeness we also mention
the scores reported in the original papers. In this table, it can be observed that both baseline models need larger 
amount of depth samples in order to maintain their performance in the higher resolution. At the highest 
density of the depth samples, BP-Net achieved a similar RMSE as under standard NYUv2 conditions in the 
original paper. CompletionFormer surpasses its own performance in terms of RMSE compared to the original 
paper when provided with the highest density of depth samples. When we use the Marigold depth estimator 
and compute the shift and scale using $\mathcal{M}$ providing 13620 depth samples, the result is far 
from the baseline models. When we steer Marigold using the same amount of points, we can see that 
the performance gets significantly closer to the baseline. All results are reported for depth sample 
neighborhood defined by $\zeta = 7$. Similarly to \cite{nair_steered_2023} we set the steering factor
$\lambda$ as a scaled constant depending on $\sqrt{1 - \bar{\alpha}_t}$ in each time step $t$.

Performance significantly drops when we erase depth samples from the area corresponding to the
medium area of $408\times248$ pixels visualized in Fig.~\ref{fig:evaluation-areas}. The depth
points are erased from the full sampling, meaning that the models are overall presented with 
fewer depth points. This scenario models the situation depicted in Fig.~\ref{fig:teaser}.
The results are presented in Table \ref{tab:best-results-missing-area}. Independently of 
the depth sparsity level, the baseline models are not able to predict depth in the erased area,
which is reflected by the worse scores in all evaluation areas (see Fig.~\ref{fig:evaluation-areas}). 
The performance of Marigold is consistent in all three evaluation areas 
due to the fact that the depth samples are used only for shift and scaling. This effectively means
that the limited amount of depth samples provided is sufficient to transform relative depth to metric 
for the entire scene. SteeredMarigold outperforms Marigold when evaluated on all three 
evaluation areas. It can be observed that SteeredMarigold achieves better score compared
to Marigold even in the areas where no depth samples are available. The score gets slightly
worse in the smallest evaluation area (see Fig.~\ref{fig:evaluation-areas}).

Finally, Fig.~\ref{fig:full-depth-vs-ma} presents depth predictions by the two considered 
baseline models and SteeredMarigold and allows for visual comparison. The area with removed 
depth samples (the second row of Fig.~\ref{fig:full-depth-vs-ma}) can be clearly observed 
as problematic in the prediction of BP-Net as well as CompletionFormer. In contrast, 
SteeredMarigold completes the scene despite the largely incomplete depth map.

\section{Discussion}
The results demonstrate that our method, SteeredMarigold, achieves competitive results compared to the
chosen top-performing baseline models in the higher resolution and increased depth sampling. SteeredMarigold
outperforms the baseline models in the scenarios reproducing the largely incomplete depth maps of
Fig.~\ref{fig:teaser}. Our approach is able to improve scale and shift of the depth estimates by appropriately steering
the---used in the plug-and-play manner---Marigold method.

Our results demonstrate, that the steering has a positive impact not only in areas with depth 
samples available, but also in large areas with no depth samples. This is supported by the better
score achieved in the medium and small areas with erased depth samples, compared to Marigold. We believe that the marginally
better score in the medium area compared to the small area (which is its sub-set), is because the 
positive impact of steering decreases with the distance of the steered areas.

The positive impact of our steering method on the regions with no depth samples can be explained by
the ability of Marigold to progressively harmonize the affected areas with the regions that are 
being actively steered. This can be visually observed in Fig.~\ref{fig:steering-missing-area}.

The steering factor $\lambda$ must be chosen carefully. The higher values lead to stronger steering
where the diffusion model is not capable of keeping up and harmonizing the remaining scene. As shown 
in Table \ref{tab:best-results}, larger steering factors lead to better score at the cost of loss of 
details generated by the diffusion model.

\section{Conclusion}
\label{sec:conclusion}
Our work has targeted a problem in-between the well-studied tasks of depth estimation and depth completion.
By design, our proposed depth completion method builds on top of the successful depth estimator Marigold.
Inheriting the qualities of Marigold, the steering process improves its prediction by manipulating the noisy 
sample over consecutive diffusion steps.

The strength of the chosen approach is its capability
to successfully complete scenes with largely incomplete depth maps, unlike the considered baseline methods: 
CompletionFormer and BP-Net. On the other hand, our approach does not outperform those baselines
when the depth samples are uniformly distributed over the entire scene, which is to be expected as those methods were specifically designed to address exactly this scenario.
Furthermore, our method comes with no guarantees of the known
depth samples keeping the same values in the final predicted depth.

Future work will focus on addressing limitations of this work.
While our method is practically training-free, it comes with a large run-time computational
cost, which prevents it from being deployable in cases where real-time performance
is desired. Eliminating the need of running the encoder and decoder in each diffusion step would allow for lighter implementations. 
Finally, considering that Marigold was trained on synthetic data only (namely on \cite{roberts:2021, cabon2020vkitti2, gaidon2016virtual})
and claims zero-shot capability, our steering method should be also evaluated on other datasets.
An obvious dataset candidate would be KITTI, which we considered in this paper for setting the sparsity
levels of the steering condition.

\bibliographystyle{ieee_fullname}
\bibliography{depth-paper}

\end{document}